\documentclass[journal]{IEEEtran}

\usepackage{hyperref}
\hypersetup{
    colorlinks=true,
    linkcolor=blue,
    filecolor=magenta,      
    urlcolor=cyan,
}
\usepackage{amsmath,amsfonts}
\usepackage{algorithmic}
\usepackage{array}
\usepackage[caption=false,font=normalsize,labelfont=sf,textfont=sf]{subfig}
\usepackage{textcomp}
\usepackage{stfloats}
\usepackage{url}
\usepackage{verbatim}
\usepackage{graphicx}
\usepackage{cite}

\usepackage{float}
\usepackage{multirow}
\usepackage{authblk}
\usepackage{url}

\usepackage[ruled,vlined]{algorithm2e}
\usepackage{listings}
\usepackage{xcolor}

\definecolor{codegreen}{rgb}{0,0.6,0}

\definecolor{codegray}{gray}{0.1}
\definecolor{codepurple}{rgb}{0.58,0,0.82}
\definecolor{backcolour}{rgb}{0.95,0.95,0.92}
\definecolor{codeblue}{HTML}{31566B}

\lstdefinestyle{mystyle}{
    commentstyle=\color{codeblue},
    basicstyle=\ttfamily\footnotesize\color{codegray},
    frame=lr,
    framesep=4pt,
    framerule=0pt,
    xleftmargin=-0.2cm,
    breakatwhitespace=false,         
    numbersep=0pt,                  
    showspaces=false,                
    showstringspaces=false,
    showtabs=false,                  
    tabsize=4
}

\usepackage{xspace}
\makeatletter
\DeclareRobustCommand\onedot{\futurelet\@let@token\@onedot}
\def\@onedot{\ifx\@let@token.\else.\null\fi\xspace}

\def\eg{\emph{e.g}\onedot} 
\def\ie{\emph{i.e}\onedot}

\def\etal{\emph{et al}\onedot}
\makeatother

\newcommand{\revise}[1]{{#1}}
\begin{document}

%
\title{Learning from 2D: Contrastive Pixel-to-Point Knowledge Transfer for 3D Pretraining}
%
%
%

\author[1]{Yueh-Cheng Liu}
\author[1]{Yu-Kai Huang}
\author[2]{Hung-Yueh Chiang}
\author[1]{Hung-Ting Su}
\author[1]{Zhe-Yu Liu}
\author[1]{Chin-Tang Chen}
\author[1]{Ching-Yu Tseng}
\author[1]{Winston H. Hsu}
\affil[1]{National Taiwan University}
\affil[2]{The University of Texas at Austin}

\maketitle
\begin{abstract}
Most 3D neural networks are trained from scratch owing to the lack of large-scale labeled 3D datasets. In this paper, we present a novel 3D pretraining method by leveraging 2D networks learned from rich 2D datasets. We propose the contrastive pixel-to-point knowledge transfer to effectively utilize the 2D information by mapping the pixel-level and point-level features into the same embedding space. Due to the heterogeneous nature between 2D and 3D networks, we introduce the back-projection function to align the features between 2D and 3D to make the transfer possible. Additionally, we devise an upsampling feature projection layer to increase the spatial resolution of high-level 2D feature maps, which enables learning fine-grained 3D representations. With a pretrained 2D network, the proposed pretraining process requires no additional 2D or 3D labeled data, further alleviating the \revise{expensive} 3D data annotation cost. To the best of our knowledge, we are the first to exploit existing 2D trained weights to pretrain 3D deep neural networks. Our intensive experiments show that the 3D models pretrained with 2D knowledge boost the performances \revise{of 3D networks} across various real-world 3D downstream tasks.
\end{abstract}

\begin{IEEEkeywords}
Pretraining, 3D neural network, 3D deep learning, contrastive learning
\end{IEEEkeywords}

%
\IEEEpeerreviewmaketitle

\section{Introduction}

3D deep learning has gained a large amount of attention recently due to its wide applications, including robotics, autonomous driving, and \revise{virtual reality}. Numerous state-of-the-art 3D neural network architectures have been proposed, showing remarkable performance improvements, including point-based methods \cite{qi2017pointnet,qi2017pointnet++,wang2019dynamic}, efficient sparse 3D CNN \cite{graham20183d,choy20194d}, and hybrid point-voxel methods \cite{liu2019point,tang2020searching}. \revise{However, unlike the well-established practice that uses large-scale datasets (\eg, ImageNet~\cite{deng2009imagenet}) to pretrain 2D neural networks for various applications, most of the 3D neural networks are trained from scratch.}
Although many efforts are made to collect 3D datasets \cite{song2015sun,dai2017scannet,geiger2013vision}, the expensive labeling cost and various 3D sensing devices make it challenging to build a large-scale dataset comparable to ImageNet, resulting in supervised pretraining in 3D difficult.

\begin{figure}[t]
\centering
 \includegraphics[width=0.8\linewidth]{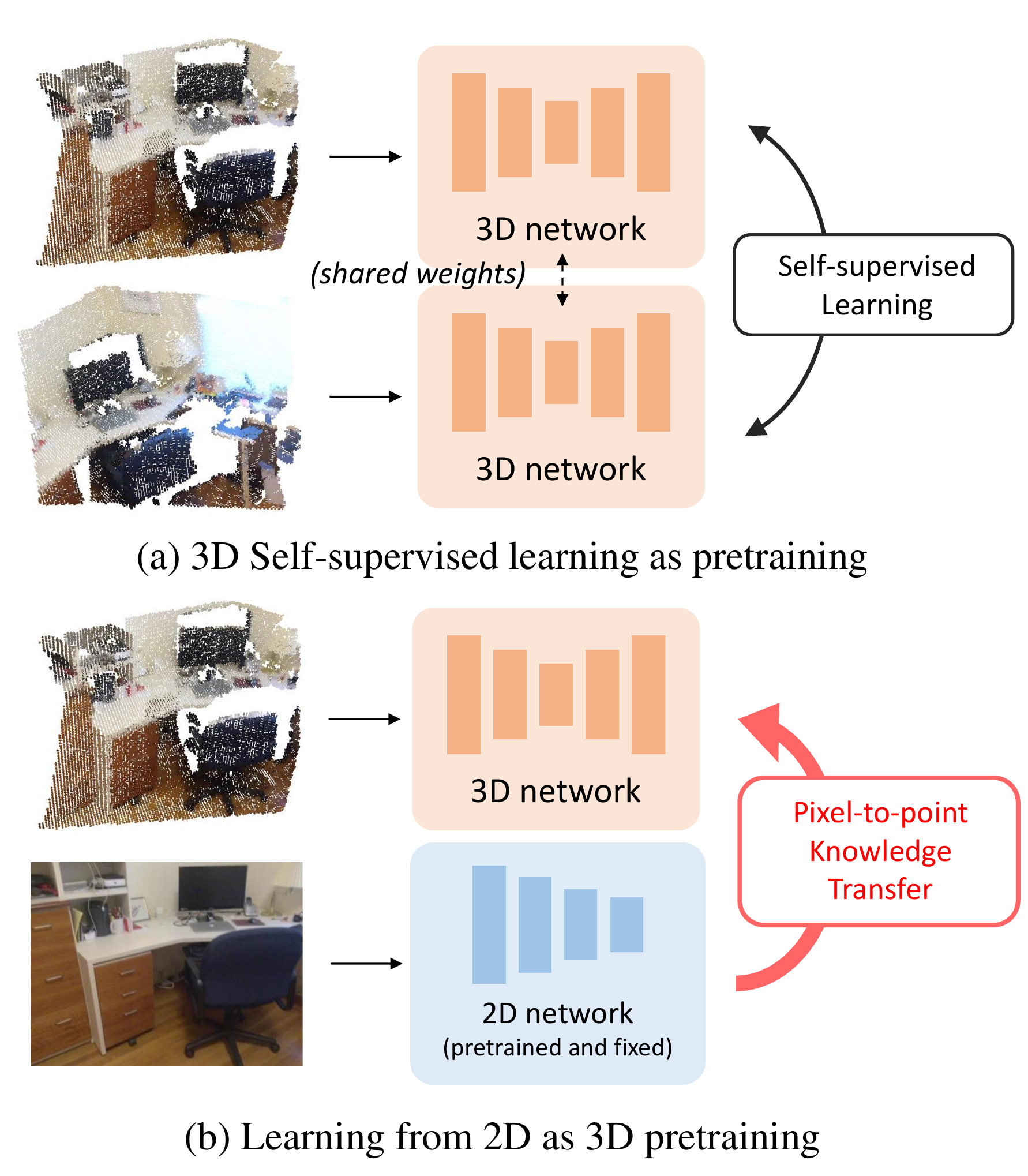}
   \caption{\textbf{Learning from 2D as 3D pretraining.} Due to the limited size of 3D labeled data, 3D pretraining with large unlabeled data is critical. (a)~Previous works \cite{xie2020pointcontrast} apply self-supervised learning on 3D as pretraining. (b)~We propose pretraining 3D networks by leveraging the existing 2D network weights with our contrastive pixel-to-point knowledge transfer. In this way, we can provide the knowledge learned from rich 2D image datasets for the 3D networks as initial model weights.}
\label{fig:illustrate}
\end{figure}

Recently, self-supervised pretraining has been proved successful in NLP \cite{devlin2018bert,radford2019language} and 2D vision \cite{he2020momentum,bachman2019learning,grill2020bootstrap,henaff2020data,hjelm2018learning,misra2020self,oord2018representation,wu2018unsupervised} \revise{by using large unlabeled data to pretrain the neural networks. This direction is promising since it can easily scale-up without the need of human annotations.} In 3D vision, PointContrast \cite{xie2020pointcontrast} first shows the opportunity of self-supervised pretraining by leveraging contrastive learning for point cloud data. \revise{By learning point correspondences across two partially-overlapping point clouds, it can contruct good initial weights for other 3D downstream tasks, such as 3D semantic segmentation and object detection.}

In this work, we study pretraining 3D neural networks for point cloud data from a novel perspective -- by learning from 2D pretrained networks (see Figure~\ref{fig:illustrate} for conceptual comparison). PointContrast \cite{xie2020pointcontrast} successfully brings the 2D self-supervised learning paradigm into 3D, but it neglects the 2D semantic information, for example, the semantic cue extracted by 2D CNN. Enormous 2D labeled datasets are much larger and more diverse than the 3D datasets, and the 2D network architectures have been widely studied and well developed in the past few years. Therefore, we believe that the knowledge of well-trained 2D networks is valuable and informative, which can provide the 3D network signals to learn good initial weights without additional labeled data.
Our idea is also related to previous 2D-3D fusion research \cite{dai20183dmv,chiang2019unified,kundu2020virtual}, suggesting that features extracted by 2D networks can complement \revise{the 3D counterparts}. The finding motivates us that learning from 2D as pretraining could bring additional information which cannot be easily learned using only 3D data.
Hence, we would like to raise a core problem: \textit{how to transfer pretrained 2D network to 3D effectively without using additional labeled data}, given the difference of their data properties and the heterogeneous network structures?

\revise{Similar to the cross-modal knowledge distillation \cite{gupta2016cross}, we view the 2D neural network as the teacher and the 3D counterpart as the student. However, unlike the 2D images and depth maps in \cite{gupta2016cross}, 2D image data and 3D point clouds are not naturally aligned. It is challenging to align intermediate network representations between 2D and 3D due to the heterogeneous network structures.}

Toward this goal, we propose the novel contrastive \textit{pixel-to-point} knowledge transfer (PPKT) for 3D pretraining. The key idea is to learn the point-level features of 3D network from the corresponding pixel representations extracted from pretrained 2D networks. \revise{Despite training the 2D network beforehand, our method does not rely on additional labeled data, saving labeling efforts for 3D data.} To enable the knowledge transfer process, we construct the pixel-point mappings and align pixel- and point-features with a differentiable back-projection function.
To overcome the lack of pixel-level outputs of common 2D networks, \eg, ResNet~\cite{he2016deep}, \revise{we introduce the learnable upsampling feature projection layer (UPL), a modified version of projection layer in common contrastive learning frameworks \cite{he2020momentum,chen2020simple}, which is applied on local image features and restores the spatial resolution of the feature map back to the original size (see Figure~\ref{fig:main})}. Our method is able to pretrain the 3D network with the knowledge of 2D networks without strict restrictions on 2D or 3D network architectures (\eg, output channel size or the output spatial resolution).

Following the pretrain-finetune protocol in \cite{xie2020pointcontrast}, we show that PPKT consistently improves overall downstream performance across multiple real-world 3D tasks and datasets. Specifically, we adopt the state-of-the-art 3D network, SR-UNet~\cite{choy20194d}, as the target network for pretraining and fine-tune on object detection and semantic segmentation tasks, including 
ScanNet~\cite{dai2017scannet}, S3DIS~\cite{armeni20163d} and SUN RGB-D~\cite{song2015sun,silberman2012indoor,janoch2013category,xiao2013sun3d}. 
We achieve +3.17 mAP@0.25 improvement on ScanNet object detection and +3.12 mIoU on S3DIS semantic segmentation compared to training from scratch. We also provide an extensive ablation study to further verify the effectiveness of our proposed method. 

Our contributions can be summarized as follows:
\revise{
\begin{itemize}
    \item We are the first to explore 3D pretraining by leveraging existing 2D pretrained knowledge for high-level, real-world 3D recognition tasks.
    \item We propose the novel contrastive pixel-to-point knowledge transfer (PPKT), enabling effective knowledge transfer from 2D to 3D as 3D pretraining.
    \item We show the effectiveness of our pretraining method by fine-tuning 3D models on various real-world 3D scene understanding tasks and successfully improve the overall performance compared to training from scratch.
\end{itemize}
}

\section{Related Work}

\subsection{3D Deep Neural Networks}
In recent years, deep neural network architectures for 3D data, \eg, point clouds or meshes, have been widely investigated. For example, MVCNN \cite{su2015multi} applies ordinary 2D convolutional neural networks (CNN) on the 2D projections of 3D objects. 3D convolutional neural networks \cite{maturana2015voxnet} perform on the quantized volumetric grid (\ie, voxels), and some specially-designed data structures and sparse convolutions \cite{riegler2017octnet,graham20183d,choy20194d} have been proposed to mitigate the high computation overhead of 3D CNN. \revise{To directly consume raw 3D point clouds, PointNet \cite{qi2017pointnet} leverages permutation invariant operators to handle the unordered point data.} Many works improve the point-based network architecture through hierarchical aggregation \cite{qi2017pointnet++,klokov2017escape,zeng20183dcontextnet}, graph-based operations \cite{xie2018attentional, verma2018feastnet, shen2018mining, chen2020hapgn}, or extending the spatial convolution on 3D point cloud \cite{hua2018pointwise, xu2018spidercnn, li2018pointcnn, su2018splatnet, thomas2019kpconv}.

\revise{
    On the other hand, some works design network architectures that fuse 2D image information with 3D data for 3D recognition tasks, such as 3D semantic segmentation \cite{dai20183dmv,chiang2019unified}, 3D object detection \cite{chen2017multi,ku2018joint,qi2020imvotenet}, and 3D instance segmentation \cite{hou20193d}. The advantage of fusing multi-modal data for neural networks is that networks for different modalities capture complementary information \cite{dai20183dmv,chiang2019unified}.
}



\subsection{Self-supervised Representation Learning}

The recent success of BERT \cite{devlin2018bert} and pretraining in NLP have brought the interest of self-supervised representation learning into computer vision. Previous methods \cite{zhang2016colorful,zhang2017split,pathak2016context,doersch2015unsupervised,noroozi2016unsupervised,gidaris2018unsupervised} train 2D neural networks by designing handcrafted self-supervised pretext tasks to learn informative representations. Recently, many works \cite{he2020momentum,bachman2019learning,chen2020simple,grill2020bootstrap,henaff2020data,hjelm2018learning,misra2020self,oord2018representation,wu2018unsupervised} utilized contrastive learning and its variants for self-supervised representation learning, which largely improve the overall performance. \revise{They also show that models learned by self-supervised contrastive learning can be used as a good initial weight for other downstream tasks.}

Some further extend the self-supervised representation learning to the 3D vision \cite{sauder2019self,rao2020global,hassani2019unsupervised}. Yet, these methods take single, clean 3D objects as input, which are not applicable to complex real-world 3D data. On the other hand, PointContrast \cite{xie2020pointcontrast} introduces contrastive-based loss designed for point clouds for 3D pretraining, especially for 3D scene understanding tasks.



\subsection{Knowledge Distillation and Cross-modal Pretraining}


Knowledge distillation (KD) or knowledge transfer \cite{bucilua2006model, li2014learning, hinton2015distilling} is proposed to compress a large model (teacher) into a smaller one (student) by learning the class-level soft-output of the teacher.
Many works enhance KD by mimicking the teacher's intermediate representations \cite{romero2014fitnets, zagoruyko2016paying}, advanced objective function \cite{huang2017like, ahn2019variational, park2019relational, peng2019correlation,tian2019contrastiverep}, or learnable proxy layers \cite{kim2018paraphrasing}. In addition to model compression, some recent works \cite{yim2017gift, zhang2018deep, furlanello2018born} utilize KD as a tool to increase the original performance of the  network.


\revise{
Knowledge distillation across modalities takes advantage of modality with rich labeled data to help the target modality with data shortage. Gupta \etal~\cite{gupta2016cross} first propose the idea that transfers the supervision of CNN from RGB images to depth maps with unlabeled pair data. It shows that pretraining with cross-modal KD can improve the performance of the model of the target modality compared to the random initialization. Similarly, some consider distillation from RGB images to sound~\cite{aytar2016soundnet} and video~\cite{diba2018spatio,girdhar2019distinit}. Note that, unlike general knowledge distillation, the distillation process here relies on paired but unlabeled data. Therefore, the cross-entropy loss that originally applies to the student and the true label cannot be used.
}

\revise{
    Despite the cross-modal knowledge distillation, many works leverage the ImageNet-pretrained CNN to initialize the neural networks on the target domain using different approaches. For example, Zhang \etal \cite{zhang2019hyperspectral} pretrain 3D neural networks by inflating the 2D image data into 3D to build a 3D pretraining dataset; Merino \etal \cite{merino20213d} transform weights of 2D convolutions into 3D convolutions as the weight initialization for the 3D neural network.
}


\section{Learning from 2D for 3D Pretraining}
\subsection{\revise{Motivation: 2D Pretraining on 3D Tasks is Effective}} \label{sec:pilot}
In the beginning, we conduct a simple pilot study showing the motivation and opportunity of pretraining deep neural networks for 3D tasks. We train a 2D semantic segmentation network to perform 3D semantic segmentation on ScanNet \cite{dai2017scannet} by aggregating multi-view predictions  \cite{chiang2019unified}. The result is shown in Table~\ref{tab:scannet-2d3d}. The 2D network with ImageNet pretraining achieves +5.75\% 2D mIoU and +4.81\% 3D mIoU improvements compared to training from scratch. The result indicates (1) \textbf{pretraining effect}: with proper 3D pretraining, 3D neural networks may perform better; (2) \textbf{2D knowledge}: 2D pretraining knowledge, \eg, ImageNet, may be helpful for 3D scene understanding tasks such as ScanNet semantic segmentation. Based on the observations, instead of developing a pure 3D self-supervised pretraining method, we aim to design a novel approach to transfer the 2D pretrained network knowledge into 3D as pretraining for 3D.

\begin{table}[h]
\begin{center}
\caption{Evaluation of ImageNet pretrained 2D FCN on SCANNET 3D semantic segmentation.}\label{tab:scannet-2d3d}
\begin{tabular}{l|c|c}
\hline
Method & 2D mIoU & 3D mIoU \\
\hline
From scratch         & 45.38 & 40.46 \\
ImageNet pretrained & 51.13 & 45.27 \\
\hline
\end{tabular}
\end{center}

\end{table}

\subsection{Contrastive Pixel-to-point Knowledge Transfer}\label{sec:ppkt}
%
\revise{
Assuming that a 2D neural network has been pretrained by a large-scale 2D image dataset (\eg, ImageNet), we aim to learn a general initial weight for the 3D network using unlabeled datasets. Then, during the fine-tune stage, the 3D network can be fine-tuned in a supervised manner for various 3D downstream tasks, such as 3D semantic segmentation or 3D object detection.
}

\begin{figure}[t]
\begin{center}
 \includegraphics[width=\linewidth]{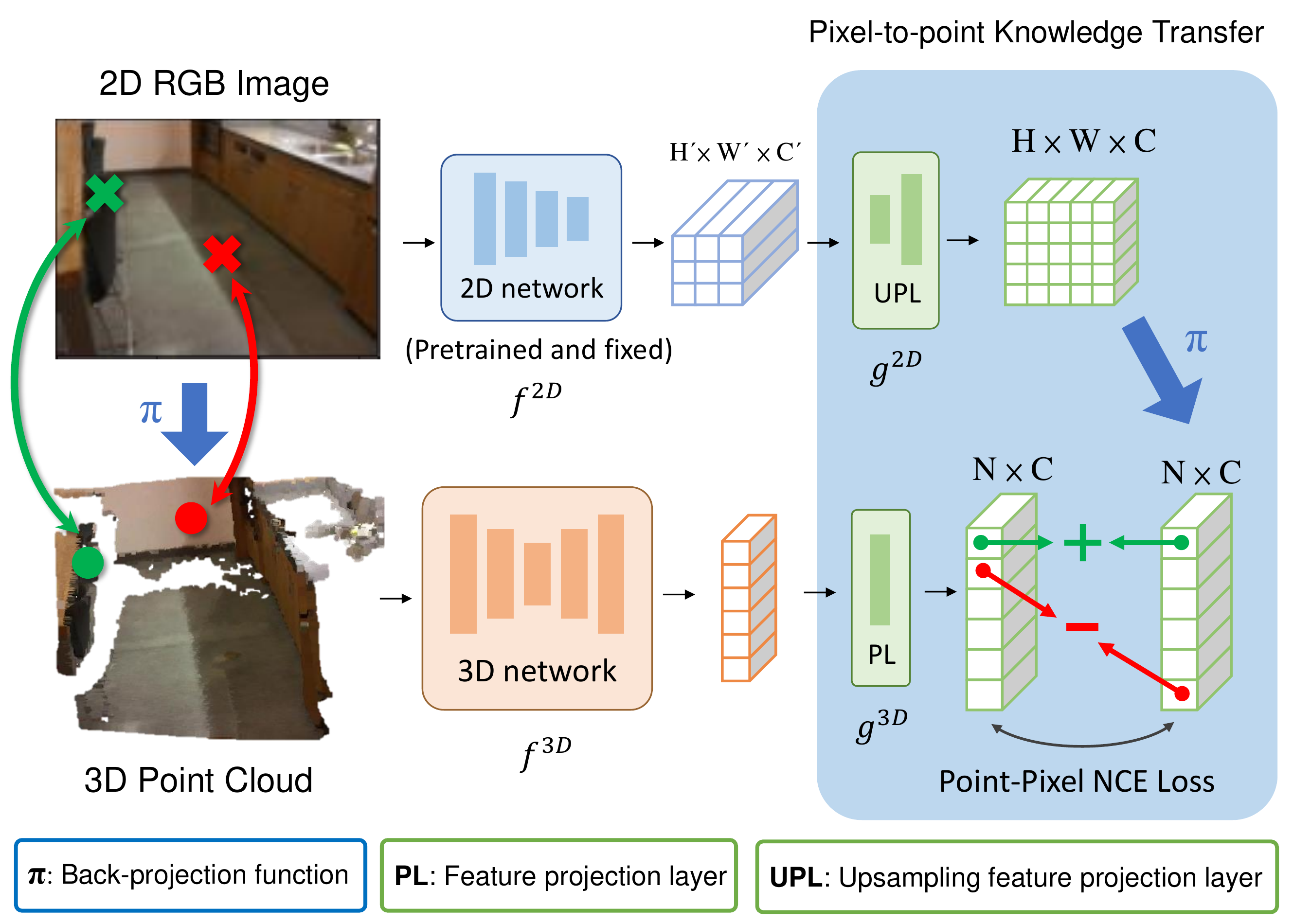}
\end{center}
   \caption{
   \textbf{Contrastive pixel-to-point knowledge transfer (PPKT).} PPKT transfers the 2D pretrained network knowledge into 3D from pixels to points.
   A back-projection is used to align corresponding pixel-level features and point-level features.
   To restore the granularity of low-resolution 2D feature maps, we propose the learnable upsampling feature projection layer (UPL). The details are described in Section~\ref{sec:ppkt}.
   }
\label{fig:main}
\end{figure}

To better exploit the 2D knowledge for 3D pretraining, we propose the novel contrastive pixel-to-point knowledge transfer (PPKT) from 2D to 3D. We illustrate the overview of our proposed method in Figure~\ref{fig:main}. For explanation, we divide our method into three parts: (1) the pixel-to-point design, (2) the upsampling feature projection layer, and (3) point-pixel NCE loss. 

\revise{
For our PPKT pretraining, we utilize a large unlabeled RGB-D image dataset to build  2D-3D data pairs. This is because unlabeled RGB-D images are easy to collect by using inexpensive RGB-D cameras, such as RealSense and Kinect.
}
\revise{
The RGB-D dataset is denoted as $\mathcal{D} = \{(\mathbf{x}_i, \mathbf{d}_i)\}^{|\mathcal{D}|}_{i=1}$, where $\mathbf{x}\in \mathbb{R}^{H \times W \times 3}$ and $\mathbf{d}\in \mathbb{R}^{H \times W}$ are the aligned RGB image and depth map.}

\revise{
Given the camera intrinsic parameters, we use the back-projection function to generate a single-view point cloud from an RGB-D image and construct the 2D-3D pixel-point mapping.
Specifically, the function is defined as $\pi\colon (\mathbf{z}, \mathbf{d}) \to (\mathbf{c}, \mathbf{f})$, where $\mathbf{z}$ is a 3-dimensional matrix with shape $H \times W \times C$, which can be an RGB image ($C=3$) or a 2D feature map ($C$ is the feature dimension); $\mathbf{c}\in \mathbb{R}^{N\times 3}$ is the coordinate of 3D points, and $\mathbf{f}\in \mathbb{R}^{N\times C}$ is the point features. In other words, $\pi$ generates a point cloud with RGB values if the input is a pair of a depth map and an RGB image. If the input is a depth map and a 2D feature map, the output point cloud has point features copied from the corresponding pixel features.
}

\revise{Additionally, we define the 2D and 3D neural networks. Let $f^{2D}$ be a 2D convolutional neural network which takes an image $\mathbf{x}$ as input and outputs a feature map with size $H' \times W' \times C$ \footnote{Without loss of generality, we ignore the final global pooling and classification layer.}. We define a 3D neural network $f^{3D}\colon (\mathbf{c}, \mathbf{f}) \to (\mathbf{c}, \mathbf{f}')$ which outputs features for each point. This formulation can represent most of the common 3D network backbones, \eg, SR-UNet~\cite{choy20194d} or PointNet++~\cite{qi2017pointnet++}.
}

\subsubsection{Knowledge transfer from pixels to points}

\begin{figure}[t]
\begin{center}
 \includegraphics[width=0.6\linewidth]{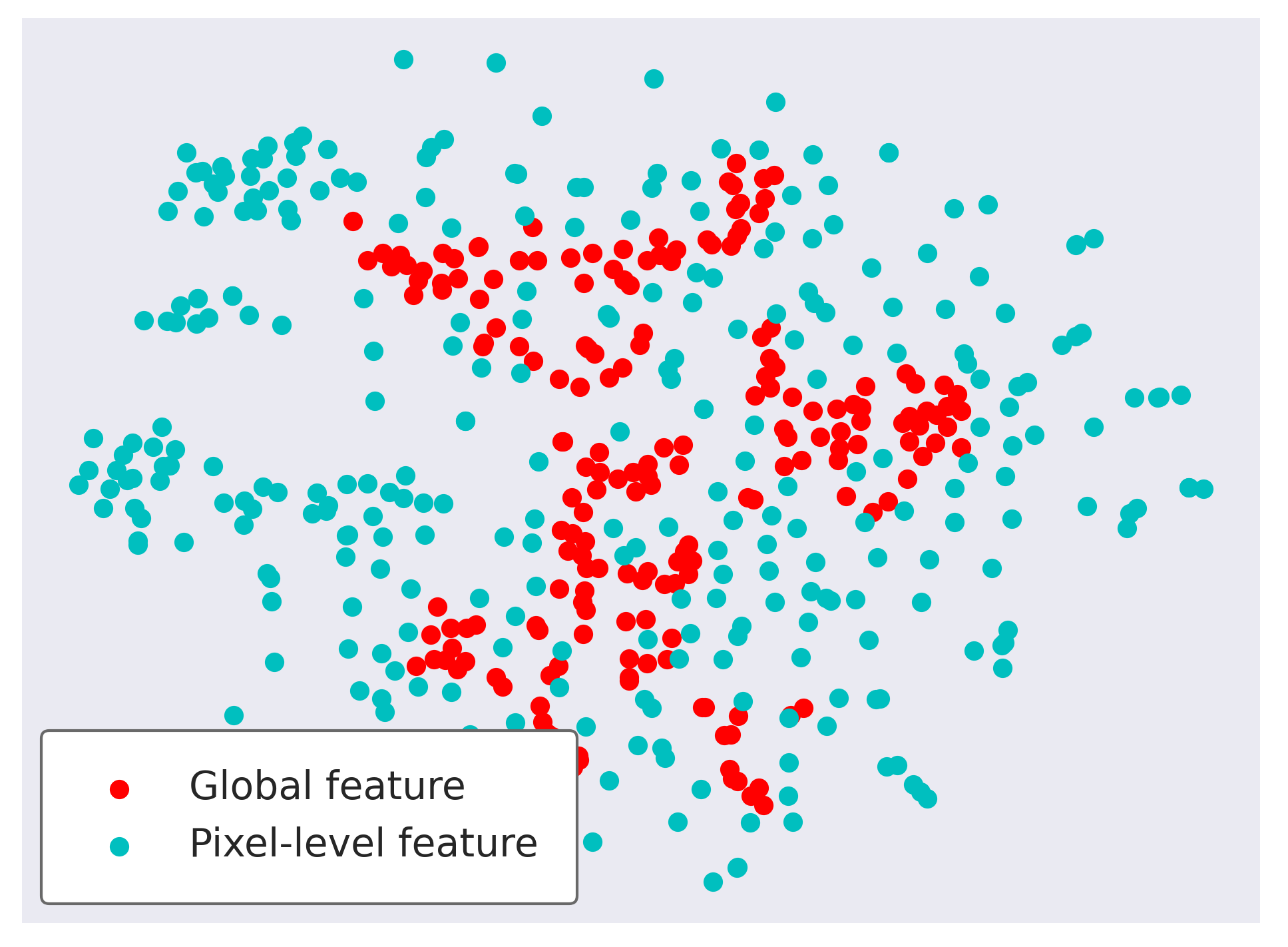}
\end{center}
   \caption{
   \revise{
   \textbf{T-SNE plot of global features versus pixel-level features.} For the images in ScanNet scene0000\_00, the global features of images extracted from ImageNet pretrained ResNet are less discriminative in feature space than pixel-level features. Therefore, we believe that pixel-level features are more favorable for knowledge transfer and can  
   preserve fine-grained pixel-level information.}
   }
\label{fig:feature}
\end{figure}
A naive cross-modal knowledge distillation approach is to minimize the distance between global features \cite{gupta2016cross}, or classification logits \cite{hinton2015distilling} of two models with different modalities. However, we found that it does not work well between 2D images and 3D point clouds in practice.
PointContrast \cite{xie2020pointcontrast} argues that 3D datasets has large number of points but small number of instances (\ie, scenes). Therefore, learning global representations for 3D will suffer from the limited number of instances.

In addition, we suppose that there are a few other reasons: (1) Global pooling operations cause spatial information loss. (2) For common 3D encoder-decoder network backbones, only the encoder would be pretrained in this way, and the decoder is ignored. (3) Unlike ordinary 2D datasets for self-supervised learning, RGB-D frames in indoor environments may contain similar global contexts, which are not discriminative in feature space. For example, most of the image frames contain the floor, walls, or tables, sharing similar global semantic meaning.
\revise{To depict our hypothesis, Figure~\ref{fig:feature} visualizes the global features and pixel-level features encoded by ImageNet pretrained ResNet~\cite{he2016deep} from indoor scene images of ScanNet dataset \cite{dai2017scannet}. The pixel-level features are more diverse in the embedding space and thus favorable for knowledge distillation.
Therefore, we propose knowledge transfer between the \textit{pixel-level} of images and the \textit{point-level} of point clouds.}

Toward this goal, we obtain the 2D and 3D feature representations encoded by the networks and align the corresponding pixels and points. Given a pair of the RGB image $\mathbf{x}$ and the depth map $\mathbf{d}$, the 3D feature representation $\mathbf{z}^{3D}$ and the pixel-to-point 2D representation $\mathbf{z}^{2D}$ are defined as 
\begin{align}
    \mathbf{z}^{3D} &= g^{3D} \Big( f^{3D}(\pi(\mathbf{x}, \mathbf{d})) \Big), \\
    \mathbf{z}^{2D} &= \pi \Big( g^{2D} ( f^{2D}(\mathbf{x})), \mathbf{d} \Big),
\end{align}
where $g^{2D}$ and $g^{3D}$ are learnable feature projection layers which map the feature of 2D and 3D into the same embedding space with identical dimension sizes. In practice, $g^{2D}$ could be a $1 \times 1$ convolution layer, and $g^{3D}$ is a shared linear perceptron. Both $g^{2D}$ and $g^{3D}$ are followed with the L2 normalization. On the other hand, $z^{2D}$ is the 2D feature representation that is back-projected into 3D space by $\pi$. In this way, $z^{2D}$ and $z^{3D}$ are well-aligned such that the 2D feature in $i$-th pixel $z^{2D}_i$ and the $i$-th point feature $z^{3D}_i$ are from the same coordinate in the 3D world.

\subsubsection{Upsampling feature projection layer (UPL)} 
\revise{
ImageNet classification pretrained network weights are commonly used in 2D computer vision since the dataset is diverse and large in scale, providing good generalizability and transferability.
However, the low spatial resolution of the feature map of the classification network (\ie, $H' \ll H$ and $W' \ll W$) makes it difficult for our contrastive pixel-to-point knowledge transfer.
}

\revise{
Therefore, we introduce the upsampling feature projection layer for $g^{2D}$ to tackle this issue. Specifically, given the feature map from the last layer of 2D CNN, we apply an $1 \times 1$ convolution and a bilinear upsampling to the original input image resolution, inspired by the decoder head in FCN \cite{long2015fully}. It also maps the 2D features into the same dimension as 3D. The method is simple yet effective. More importantly, it enables the flexibility to handle the spatial resolution differences and channel sizes  of various 2D network architectures.
}

\subsubsection{Point-pixel NCE loss} As the learning objective, we minimize the relative distance between the corresponding pixel and point representation through point-pixel NCE loss (PPNCE), a modified version of InfoNCE loss \cite{oord2018representation} for contrastive learning, which is defined as
\begin{equation}
    \mathcal{L}_{\text{PPNCE}}
    = -\sum_{i=1}^{N} \log \frac{
        \exp (\mathbf{z}^{3D}_i \cdot \mathbf{z}^{2D}_i / \tau)
    }{
        \sum_{j=1}^{N} { \exp (\mathbf{z}^{3D}_i \cdot \mathbf{z}^{2D}_j / \tau)}
    }    
\end{equation}
where $\tau$ is the temperature hyper-parameter, \revise{the $\cdot$ notation here is the dot-product, and $N$ is the total number of points. The physical meaning of the loss function is to form a feature space by attracting a 3D point feature and its corresponding 2D pixel feature while separating the 3D feature from other 2D features at the same time. In other words, if a pixel and a point share the same coordinate in the 3D world, they are positive pairs; otherwise, they are negative pairs.} For empirical memory concern, we sub-sample a fixed number of points and pixels for the loss calculation. 

Note that most existing contrastive learning methods apply linear or non-linear feature projection layers on the \textit{global} features by multi-layer perceptrons \cite{he2020momentum,chen2020simple,tian2019contrastiverep}. In contrast, our 3D feature projection layer $g^{3D}$ is applied on 3D point-level feature maps, and the 2D upsampling projection layer $g^{2D}$ is applied on 2D high-level but low-resolution feature maps.
For simplicity, we do not include memory bank \cite{chen2020simple} due to the large number of pixels and points.

\subsubsection{Discussion}
\revise{
Our problem formulation is similar to the cross-modal knowledge distillation \cite{gupta2016cross}, considering that the 2D network is the teacher and the 3D network is the student. However, the key difference is that that RGB images and depth maps in \cite{gupta2016cross} are well-aligned while 2D images and 3D point cloud are not, and the networks for 2D and 3D are heterogeneous.
}

\revise{
If we consider the back-projected pixel-level features as pseudo 3D points, our loss function is similar to PointInfoNCE~\cite{xie2020pointcontrast}. We both apply contrastive loss on point-level (or pixel-level) instead of global instance level, which is commonly presented in 2D contrastive learning. However, our PPNCE is applied across 2D and 3D features encoded by two \textit{different networks} from the same data pair. In contrast, PointInfoNCE is applied across features of two different point cloud samples extracted by the \textit{same 3D network}. Additionally, our motivation is different. We propose using contrastive loss to minimize the relative distance between the point and the pixel representation to transfer the 2D knowledge to 3D as pretraining.
}


\section{Experiment}
\revise{
In this section, we study the effectiveness of our proposed 3D pretraining method. The experiment focuses on the ability to transfer the 3D network pretrained by our method to various downstream tasks and whether it can improve the performance compared to training from scratch. Specifically, the evaluation is divided into three steps: (1) Given a pretrained 2D neural network, we pretrain the 3D neural networks by PPKT with an unlabeled RGB-D dataset. (2) We fine-tune the model for the target downstream tasks in a supervised manner, including 3D semantic segmentation and 3D object detection. (3) We evaluate the fine-tune performance on the testing data.
In the following, we will describe the details and results.
}

\subsection{Experimental Setup of 3D Pretraining}

\subsubsection{Network architectures}
\revise{
Our PPKT requires a pretrained 2D neural network as the teacher. Therefore, we choose the most widely-used ResNet \cite{he2016deep} pretrained on ImageNet classification as our 2D backbone. For the 3D network, following \cite{xie2020pointcontrast}, we adopt the Sparse Residual 3D U-Net 34 (SR-UNet34) \cite{graham20183d,choy20194d} since it achieves state-of-the-art performance on multiple 3D tasks.} It contains 34 layers of 3D sparse convolution layers with the encoder-decoder design and skip connections. SR-UNet is innately suitable for the 3D semantic segmentation task since it generates per-point outputs. Following \cite{xie2020pointcontrast}, it can also perform 3D object detection by attaching VoteNet \cite{qi2019deep} modules.

\subsubsection{Datasets}
\revise{
For pretraining, we use the raw RGB-D images in ScanNet dataset \cite{dai2017scannet}, collected by hand-hold light-weight depth sensors, since it is currently the largest available real-world dataset of its kind.} ScanNet contains 1513 indoor scans for 707 distinct spaces. We sub-sample every 25 frames from the sequential RGB-D data, resulting in about 100k frames in total.

\subsubsection{Baselines} 
\revise{
We consider different pretraining approaches for the 3D neural network as baselines. In particular, we compared our method with PointContrast~\cite{xie2020pointcontrast} using the officially-released pretrained weight. 
}
Additionally, to prove the effectiveness of our design, we build other naive 2D-3D knowledge transfer methods for comparison. Global knowledge distillation (Global KD) represents the method we modified from \cite{hinton2015distilling}, considering the SR-UNet \textit{encoder} as the student and 2D CNN as the teacher. Precisely, we extract the encoder part from SR-UNet and attach a mean pooling layer and a classifier where the output number of classes matched with the ImageNet supervised pretrained network (\ie, output 1000 classes). We minimize the KL-divergence between 2D and 3D logits. Even though the dataset we used is not from ImageNet, we force the 2D classifier to output ``dark'' class information as the learning target for the 3D network. Note that since the dataset has no label, the classification loss between student and true label is ignored. In other words, the student (SR-UNet encoder) is purely mimicking the teacher's output.

\revise{
CRD \cite{tian2019contrastive} follows the similar settings with Global KD but minimizes the contrastive-based loss function instead of traditional KD loss. We also compared with \cite{gupta2016cross} with slight modifications to adapt to our 2D-3D settings, which is minimizing the L2 distance between 2D and 3D global features. Note that for Global KD, CRD, and \cite{gupta2016cross} baselines, only the 3D encoder is pretrained. We provide the summary of baselines in Table~\ref{tab:baselines}.}

\begin{table}[h]
\centering
\caption{The summary of the 3D pretraining baselines}
\label{tab:baselines}
\setlength\tabcolsep{2pt}
\begin{tabular}{llll}
\hline
Method        &  Objective                    & Pretrained 3D network & Loss function           \\
\hline
Global KD     & 2D $\to$ 3D            & encoder         & Global KL-div    \\
CRD     \cite{tian2019contrastiverep}      & 2D $\to$ 3D            & encoder         & Global contrastive \\
Gupta \etal  \cite{gupta2016cross}       & 2D $\to$ 3D            & encoder         & Global L2          \\
PointContrast \cite{xie2020pointcontrast} & 3D $\leftrightarrow$ 3D & encoder-decoder & Local contrastive  \\
PPKT (ours)   & 2D $\to$ 3D            & encoder-decoder & Local contrastive \\
\hline
\end{tabular}
\end{table}

\subsubsection{Training details of PPKT} 

We apply the upsampling feature projection layer on the 2D ResNet50 layer4 output. As for 3D, a feature projection layer is used to project the last layer feature of SR-UNet. The projected feature dimension 128. The temperature of NCE loss is 0.04. The voxel size is set as 2.5cm, and the input image size is [480, 640]. We use momentum SGD with learning rate 0.5 and weight decay 1e-4. The exponential learning rate scheduler is used. We apply 2D augmentations, including horizontal flip and random resize, and 3D augmentations, including scaling, rotation, and elastic distortion. We pretrain the 3D network for 60k iterations using one V100 GPU with batch size 24. \revise{We use the same PPKT pretrained 3D network as the initial weight for all the downstream tasks and datasets during fine-tune stage.}

\subsection{Fine-tune on S3DIS Semantic Segmentation}

\subsubsection{Setup} Stanford Large-Scale 3D Indoor Spaces dataset (S3DIS) \cite{armeni20163d} contains 6 large buildings, nearly 250 rooms, and semantic labels of 13 categories. For evaluation, we follow the most common setup which takes area 5 for validation. For training, we use 5cm voxel size and follow most of the details in \cite{xie2020pointcontrast}. We use SGD+momentum with learning rate 0.1 and polynomial learning rate scheduler. We apply common 3D data augmentations during fine-tuning, including scaling, rotation, chromatic distortion, elastic distortion, translation, and point dropout. We use batch size 32 and one V100 GPU training for 15k iterations. \revise{Note that we apply the same fine-tune details for all the baselines' and our pretrained 3D models, including the random seed.} \revise{The evaluation metrics are the mean class accuracy and mean class IoU.} We report the final validation performance.

\begin{table*}[!thb]
    \begin{minipage}{.5\linewidth}
        \centering
        \caption{Fine-tune result on S3DIS semantic segmentation.}
        \label{tab:s3dis}
        \begin{tabular}{lcc}
        \hline
        Method         & mean Acc & mean IoU \\
        \hline
        From scratch    & 73.24 & 65.16     \\ \hline
        Global KD       & 72.65 & 66.56 \\
        Gupta \etal \cite{gupta2016cross} & 72.11 & 64.39 \\
        CRD \cite{tian2019contrastiverep} & 72.70 & 65.65 \\ 
        PointContrast \cite{xie2020pointcontrast} & 73.97 & 66.86  \\
        PPKT (ours) & \textbf{75.19} & \textbf{68.28} \\
        \hline
        \end{tabular}
    \end{minipage}%
    \begin{minipage}{.5\linewidth}
        \centering
        \caption{Fine-tune result on SUN RGB-D 3D object detection.}
        \label{tab:sunrgbd}
        \begin{tabular}{lcc}
        \hline
        Method         & mAP@0.25  & mAP@0.5 \\
        \hline
        From scratch    & 55.21     & 32.81    \\\hline
        Global KD & 56.10 & 33.58 \\
        Gupta \etal \cite{gupta2016cross} & 54.42 & 31.12 \\
        CRD \cite{tian2019contrastiverep} & 55.86 & \textbf{34.30} \\ 
        PointContrast \cite{xie2020pointcontrast} & 56.14     & 32.70 \\
        PPKT (ours) & \textbf{57.26} & 33.92 \\
        \hline
        \end{tabular}
    \end{minipage} 
\end{table*}

\subsubsection{Result} The result of S3DIS semantic segmentation is shown in Table~\ref{tab:s3dis}. Compared to training from scratch, our proposed contrastive pixel-to-point knowledge transfer pretraining brings significant improvements (+3.34\% mIoU). \revise{In contrast, the performances of global learning from 2D baselines (Global KD, CRD, Gupta \etal) are similar to training from scratch, which indicate that the pretraining effects of these baselines are limited. The comparison suggests that (1) pretraining the encoder-decoder is helpful, and (2) our pixel-to-point design is important to effectively transfer 2D knowledge since those baselines and our method all use the same ImageNet-pretrained 2D teacher network.}

\begin{table}[t]
\begin{center}
    \caption{Fine-tune result on ScanNet semantic segmentation and object detection. See Section~\ref{sec:scannet-exp} and Figure~\ref{fig:scannet-limited} for more detail on semantic segmentation result.}
    \label{tab:scannet-full}
    \begin{tabular}{lcccc}
    \hline
     & \multicolumn{2}{c}{Semantic segmentation} &  \multicolumn{2}{c}{Object detection} \\
    Method        & mean Acc & mean IoU  & mAP@0.25  & mAP@0.5 \\
    \hline
    From scratch    & 88.57 & 69.49 & 56.50     & 34.54    \\\hline
    Global KD & \textbf{88.63} & \textbf{70.35} & 57.50 & 34.81 \\
    Gupta \etal \cite{gupta2016cross} & 88.40 & 68.71 & 56.74 & 36.27 \\
    CRD \cite{tian2019contrastiverep} &  88.08 & 68.53 & 56.82 & 36.75 \\ 
    PointContrast \cite{xie2020pointcontrast} & \textbf{88.63} & 69.22 & 58.30     & 36.26 \\
    PPKT (ours) & 88.53 & 69.56  & \textbf{59.67} & \textbf{38.90} \\
    \hline
    \end{tabular}
\end{center}
\end{table}
\subsection{Fine-tune on SUN RGB-D Object Detection}

\subsubsection{Setup} SUN RGB-D dataset \cite{song2015sun} comprises about 10,000 RGB-D images and is annotated with 60k 3D oriented bounding boxes of 37 object classes. The RGB-D images are back-projected into 3D point clouds. We use the official train/val split and train with the ten most common object classes. For training details, we use 64 for batch size and 5cm voxel size. We use Adam optimizer with 0.001 learning rate training for 180 epochs on one V100 GPU. The learning rate is decreased 10x after 80 and 120 epochs. For data augmentations, we adopt random rotation, scaling, and flipping. \revise{The evaluation metric is the mean AP considering two 3D IoU threshold, 0.25 and 0.5.} We report the final validation performance.



\subsubsection{Result} 
\revise{
The result of our pretrained model against training from scratch is summarized in Table~\ref{tab:sunrgbd}. 3D object detection requires high-level semantic understanding for object classification and localization. Our proposed contrastive pixel-to-point knowledge transfer achieves significant improvement compared to training from scratch (+2.28\% mAP@0.25). The performance improvement is consistence with the result in the S3DIS semantic segmentation, showing the generalization ability of our pretrained weights.
}

\subsection{Fine-tune on ScanNet Semantic Segmentation and Object Detection}\label{sec:scannet-exp}

\subsubsection{Setup} ScanNet \cite{dai2017scannet} dataset contains 1201 scans in training set and 312 in validation set. For semantic segmentation, we use the v2 label set, which has 20 semantic classes and evaluate the performance on vertex points. We use 2cm voxel size. We split the origin training set into train/val 949/525 scenes. We save the best validation score and test on the original validation set. We use SGD+momentum with learning rate 0.1 and polynomial learning rate scheduler. The batch size is 12 for a single GPU. We use 4 V100 GPUs training for 15k iterations. We use the same augmentations as in S3DIS. During testing, the points are assigned to the predicted class of the nearest voxel center.

For object detection, ScanNet contains 18 object classes for instance segmentation labeled on mesh data. We follow the preprocessing  of \cite{qi2019deep,xie2020pointcontrast}, generating 3D axis-aligned bounding box. Compared to SUN RGB-D, ScanNet consists of reconstructed 3D scans, which are larger and more complete. We set 2.5cm for voxel size and follow the training details in \cite{xie2020pointcontrast}.

\begin{figure}[t]
\centering
\includegraphics[width=0.9\linewidth]{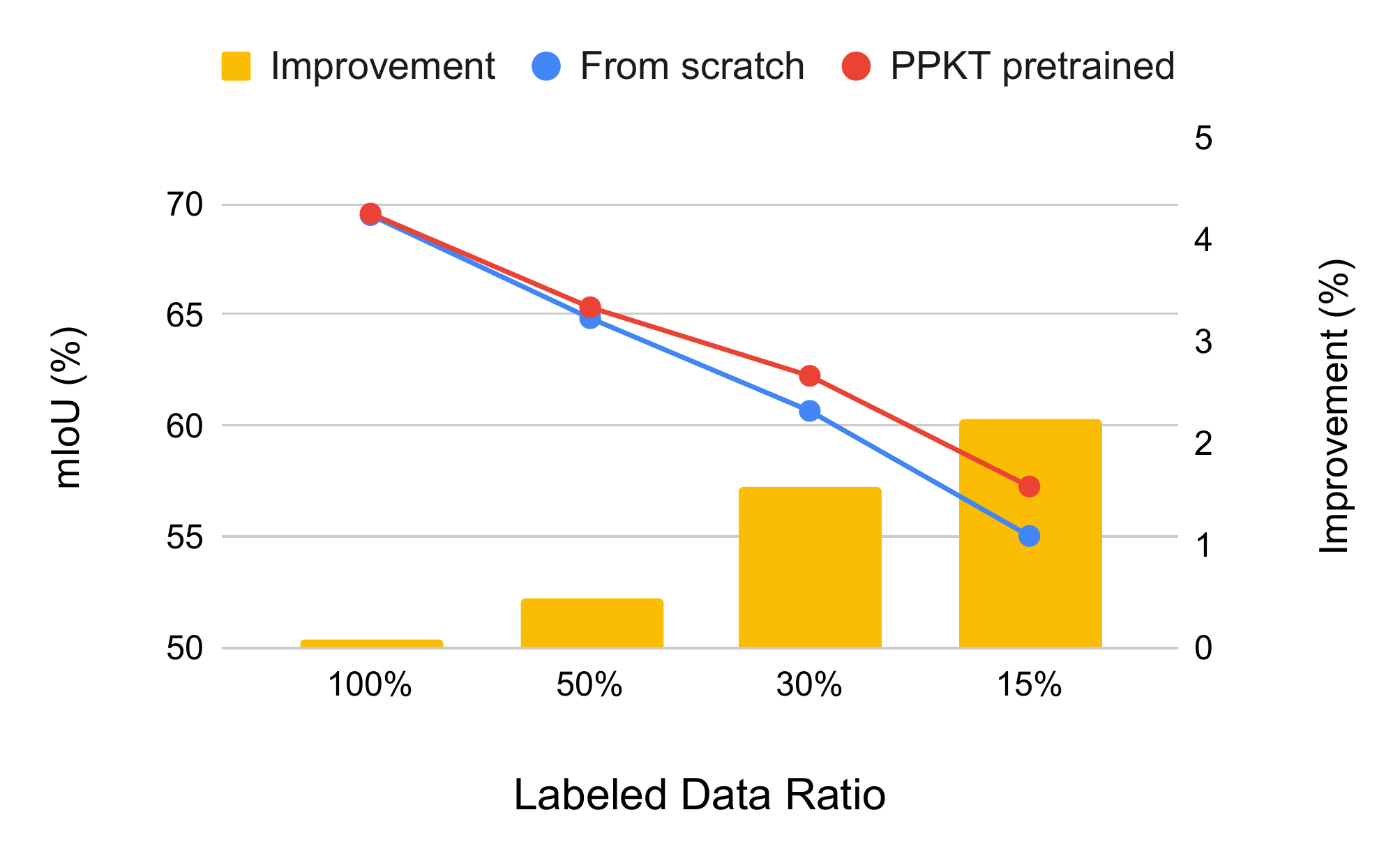}
\vspace{-2mm}
   \caption{\textbf{Limited labeled data fine-tuning on ScanNet semantic segmentation.} We sub-sample the labeled scenes in the ScanNet dataset into 50\%, 30\%, and 15\%. With less labeled data available, the gap between our PPKT pretraining and training from scratch becomes larger.}
\label{fig:scannet-limited}
\end{figure}

\subsubsection{Result} The semantic segmentation and object detection results are shown in Table~\ref{tab:scannet-full}. For semantic segmentation, the pretraining has little performance improvement against training from scratch. Since the supervised fine-tune dataset is the same as our pretraining dataset, we suggest that self-supervised pretraining only accelerates the training process but hardly brings performance improvements. For verification, we experiment reducing the number of labeled data in ScanNet for fine-tuning gradually. We randomly sample 50\%, 30\%, 15\% of the labeled scenes in the dataset for fine-tuning, and the result is presented in Figure~\ref{fig:scannet-limited}. Although there is no pretraining gain using 100\% labeled data during fine-tuning, the performance gap between pretraining and training from scratch becomes larger when the data size is smaller (2.23\% mIoU difference for 15\% labeled data). Our limited labeled data fine-tune result also matches the findings in previous works \cite{newell2020useful,chen2020big}, which suggest that self-supervised pretraining methods benefit more when the size of labeled data is smaller in the downstream supervised task.

For object detection, unlike the semantic segmentation result, our contrastive pixel-to-point knowledge transfer increases the performance by a large margin (+3.17\% mAP@0.25 and +4.36\% mAP@0.5) even if the pretraining dataset and fine-tune dataset are the same. The improvements are consistent with the results in SUN RGB-D object detection. This may suggest that 3D object detection tasks benefit more from the high-level semantics learned from the 2D pretrained knowledge.

\subsection{Ablation Study}

\subsubsection{Backbone network size}
Previous works \cite{chen2020big,newell2020useful} have shown that larger or deeper 2D networks will benefit more from self-supervised pretraining. \revise{Therefore, we would like to study the effect of our PPKT pretraining for different 3D network sizes. We provide the ablation study on S3DIS in Table~\ref{tab:backbone}.} SR-UNet18 has about half the size in terms of network parameters compared to SR-UNet34. \revise{The result shows that our PPKT pretraining on larger models has more performance improvement, which is similar to the finding in the previous works.} It is also worth noticing that our PPKT pretraining on SR-UNet18 could achieve the performance of SR-UNet34 training from scratch. In other words, a good pretraining can make up the network backbone size limitation.

\begin{table}[t]
\begin{center}
\caption{\textbf{3D backbone ablation study on S3DIS.} The models with larger capacity benefit more from pretraining with rich unlabeled data.}
\label{tab:backbone}

\begin{tabular}{lcc}
\hline
Method         & mean Acc & mean IoU \\
\hline
SR-UNet18           & 71.67 & 64.67 \\
SR-UNet18 + PPKT    & 73.67 (+2.0) & 66.40 (+1.73) \\\hline
SR-UNet34           & 73.24 & 65.16 \\
SR-UNet34 + PPKT    & \textbf{75.19} (+1.95) & \textbf{68.28} (+3.12) \\
\hline
\end{tabular}
\end{center}
\end{table}

\subsubsection{Point-pixel loss functions and datasets}\label{sec:ablation-loss}

\begin{table}[t]
\begin{center}
\caption{\textbf{Point-pixel loss ablation study on SUN RGB-D object detection.} PPNCE loss is preferred over PPKD. See Section~\ref{sec:ablation-loss} for more detail.}
\label{tab:ade20k}
\begin{tabular}{lccc}
\hline
Loss  & 2D Dataset & 2D Network & \small{mAP@0.25}  \\
\hline
From scratch  & - & -  & 55.21 \\
PPKD loss &  ADE20k & ResNet-FCN  & 56.07 \\
PPNCE loss & ADE20k & ResNet-FCN & \textbf{58.03}  \\
PPNCE loss & ImageNet & ResNet    & 57.26\\
\hline
\end{tabular}
\end{center}
\end{table}

We study the loss function in our pixel-to-point knowledge transfer by replacing PPNCE loss with the ordinary knowledge distillation loss \cite{hinton2015distilling} applied on points and pixels (PPKD). However, ImageNet pretrained ResNet does not have pixel-wise class-level predictions. To achieve this, we train a 2D ResNet-FCN \cite{long2015fully} with ADE20k semantic segmentation dataset \cite{zhou2017scene} as the 2D teacher network. 
We evaluate the fine-tune performance on SUN RGB-D object detection in Table~\ref{tab:ade20k}. The result shows that pretraining with both loss functions are better compared to training from scratch, and PPNCE is better than PPKD (+2.03\% mAP@0.25). This is because, in contrast to traditional KD loss, which treats each output independently, the contrastive loss is better to transfer the knowledge in structurally due to the large numbers of negative examples \cite{tian2019contrastiverep}.

On the other hand, ADE20k PPNCE performs better than the ImageNet PPNCE.
\revise{Nevertheless, we believe that 2D teacher network architectures without decoder (\eg, FCN) combined with the UPL are the better choices since FCN-like structure requires semantic segmentation dataset to pretrain. In contrast, our default setup has more flexibility and weaker assumption on 2D network architectures.}

\subsubsection{Learning from 2D self-supervised pretrained teacher network}\label{sec:moco}
\revise{
We further show that our PPKT is not limited to 2D teacher networks that are pretrained by supervised learning (\eg, ImageNet classification). It can also take advantage of 2D self-supervised pretrained model as teacher. Here,
we use a 2D ResNet50, which is self-supervised pretrained by MoCo \cite{he2020momentum} beforehand, as teacher for PPKT and compared to our default settings, where ImageNet supervised teacher is used. 
}

\revise{
We show the result in Table~\ref{tab:moco}. In our experiment, the 3D model learned from MoCo-pretrained 2D teacher shows comparable fine-tune performance with the 3D model learned from the supervised ImageNet classification teacher. We believe that the performance could be further improved if the 2D self-supervised teacher is pretrained with an unlabeled dataset larger than ImageNet, which is the strength of self-supervised learning.
}

\begin{table}[t]
\begin{center}
\caption{\textbf{PPKT with self-supervised pretrained 2D network as teacher on 3D object detection.} See Section~\ref{sec:moco} for more detail.}
\label{tab:moco}
\begin{tabular}{lcc}
\hline
\multirow{2}{1em}{Method}

 &  \multicolumn{2}{c}{mAP@0.25} \\
 & ScanNet & SUNRGBD \\
 \hline
From scratch   & 56.50 & 55.21 \\
PPKT (ImageNet super.) & 59.67 & \textbf{57.26} \\
PPKT (ImageNet MoCo \cite{he2020momentum}) & \textbf{59.69} &57.17 \\
\hline

\end{tabular}
\end{center}

\end{table}

\subsection{Visualization}
\begin{figure}[h]
\begin{center}
\includegraphics[width=0.8\linewidth]{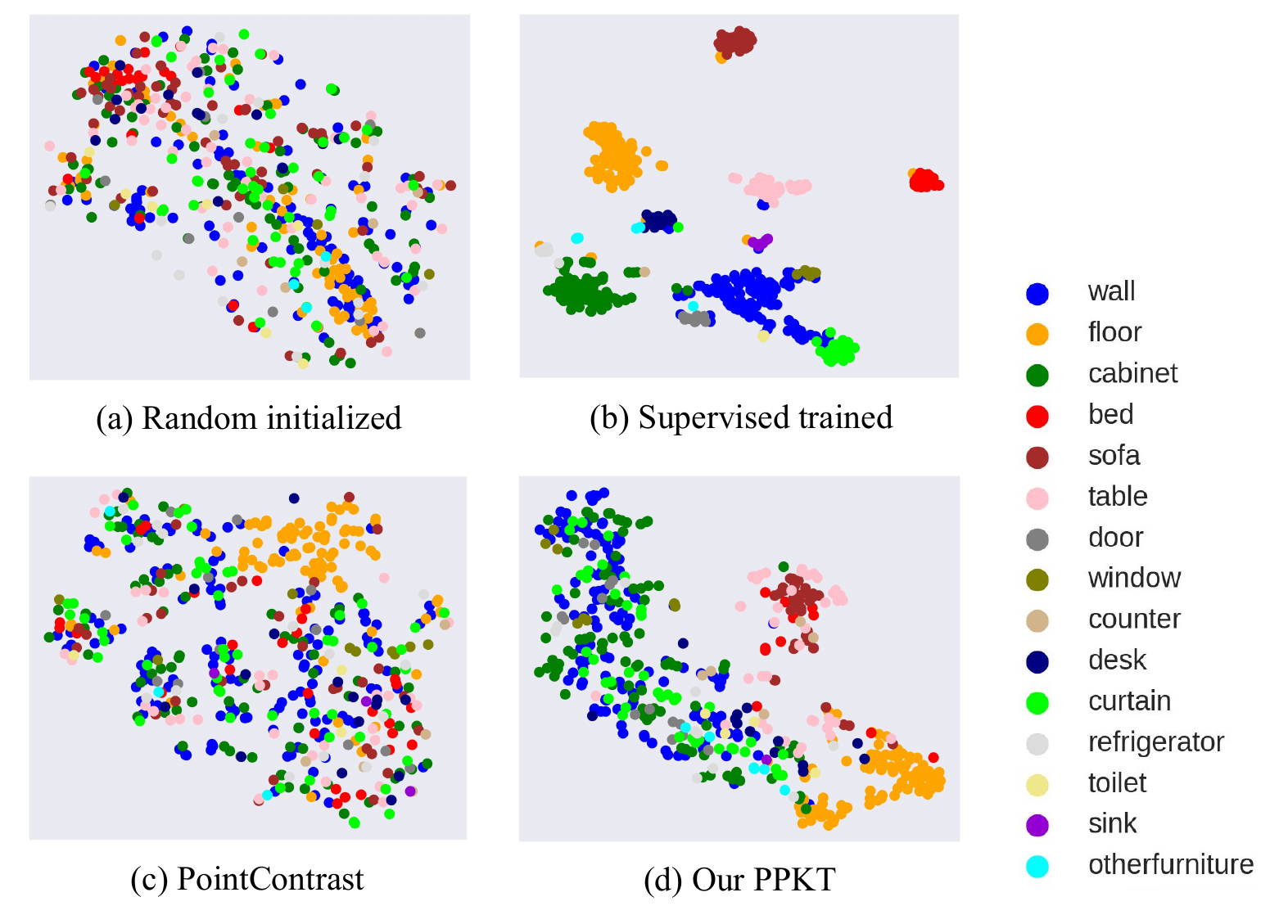}
\end{center}
   \caption{\textbf{T-SNE visualization of point-level features in ScanNet.} We visualize the last-layer point features of a 3D scene using different 3D network weights. The feature points are colored with segmentation class labels. Without any 3D labeled data, our method shows the high-level semantic understanding ability by learning from 2D rich information.}
\label{fig:visualization}
\end{figure}

\revise{
We provide the visualization of the learned features of PPKT \textit{without fine-tuning} in Figure~\ref{fig:visualization}. We extract the point-level features from the last layer of SR-UNet (before the segmentation head) of the 3D scan scene0000\_00 in the ScanNet dataset and visualize them with T-SNE. For fair comparison, except for the network weights, all the other settings remain identical. Remarkably, without any labeled data during pretraining, our pretrained network is able to encode semantic meaningful point embedding by exploiting the rich 2D pretrained knowledge. This shows that learning from 2D with PPKT is a promising approach to provide pretraining weights for 3D neural networks, especially for those tasks that require high-level semantic understanding.
}

\section{Conclusion}
\revise{
    In this work, we explore a new 3D pretraining approach by learning from 2D pretrained networks. We present the contrastive pixel-to-point knowledge transfer (PPKT) to exploit the 2D knowledge effectively. We show that combining the pixel-to-point design and the contrastive loss helps learn good initial weights for 3D models with large unlabeled RGB-D data. We also demonstrate that 3D pretraining through PPKT is more effective when the network size is larger, or the fine-tune labeled dataset is limited. Moreover, our method is complementary to PointContrast~\cite{xie2020pointcontrast}. We expect that our idea of learning from 2D and empirical findings could inspire future works to consider 2D network knowledge when developing self-supervised 3D algorithms.
}

\appendix[Experiment Result Details]

We provide per-class performance for S3DIS in Table~\ref{tab:s3dis-det}, SUN RGB-D in Table~\ref{tab:sunrgbd-det}, and ScanNet in Table~\ref{tab:scannet-det}.

\begin{table*}[h]
    \centering
    \caption{\textbf{Per-class IoU on S3DIS semantic segmentation.}}
    \label{tab:s3dis-det}

    \small
    \setlength\tabcolsep{2pt}
    \begin{tabular}{l|ccccccccccccc|cc}
    \hline
    Method & clutter & beam & board & book & ceil & chair & column & door & floor & sofa & table & wall & window & mIoU & mAcc \\ \hline
    From scratch & 56.66 & \textbf{0.03} & 72.83 & 68.68 & 92.78 & 89.17 & 31.49 & 68.03 & 98.07 & 58.64 & 76.49 & 82.83 & 51.43 & 65.16 & 73.24 \\
    PointContrast & \textbf{62.01} & 0.00 & 74.39 & \textbf{74.07} & \textbf{94.67} & 90.18 & \textbf{37.35} & 70.33 & \textbf{98.54} & 53.35 & 78.80 & \textbf{84.45} & 51.02 & 66.86 & 73.97 \\
    PPKT (ours) & 57.15 & 0.00 & \textbf{78.05} & 73.26 & 93.05 & \textbf{91.00} & 33.31 & \textbf{77.11} & 98.44 & \textbf{70.46} & \textbf{79.77} & 83.67 & \textbf{52.25} & \textbf{68.27} & \textbf{75.19} \\
    \hline
    \end{tabular}

\end{table*}

\begin{table*}[h]
    \centering
    \caption{\textbf{Per-class AP under IoU 0.5 on SUN RGB-D object detection.}}
    \label{tab:sunrgbd-det}
    \small
    \begin{tabular}{l|cccccccccc|c}
    \hline
    Method & bed & table & sofa & chair & toilet & desk & dress & night & book & bath & mAP \\ \hline
    From scratch & 47.35 & \textbf{18.58} & 47.96 & 52.23 & \textbf{61.74} & 5.31 & \textbf{15.65} & 33.16 & 5.69 & \textbf{40.44} & 32.81 \\
    PointContrast & 49.95 & 18.38 & 48.52 & \textbf{53.84} & 53.69 & 5.61 & 14.68 & 36.87 & 7.89 & 37.58 & 32.70 \\
    PPKT (ours) & \textbf{52.10} & 17.61 & \textbf{50.34} & 52.31 & 58.05 & \textbf{6.20} & 12.98 & \textbf{43.72} & \textbf{9.36} & 36.55 & \textbf{33.92} \\
    \hline
    \end{tabular}
    
\end{table*}

\begin{table*}[h]
    
    \setlength\tabcolsep{1.5pt}
    \centering
    \caption{\textbf{Per-class AP under IoU 0.5 on ScanNet object detection.}}
    \label{tab:scannet-det}
    \small
    \begin{tabular}{l|cccccccccccccccccc|c}
    \hline
    Method & cabin & bed & chair & sofa & table & door & wind & book & pic & cntr & desk & curtn & refrig & shower & toilet & sink & bath & garbg & mAP \\ \hline
    From scratch & 10.61 & 68.40 & 64.86 & 53.54 & 37.69 & 20.64 & 9.80 & 30.08 & 0.64 & 13.92 & 36.63 & 24.70 & \textbf{31.32} & 10.57 & 83.49 & 21.66 & 83.39 & 19.73 & 34.54 \\
    PointContrast & \textbf{12.41} & 67.22 & 70.06 & \textbf{57.91} & \textbf{46.54} & 22.72 & \textbf{11.59} & 37.12 & 0.30 & 10.10 & 36.35 & \textbf{25.47} & 27.30 & 21.64 & 84.96 & \textbf{23.84} & 75.30 & 21.84 & 36.26 \\
    PPKT (ours) & 7.95 & \textbf{70.25} & \textbf{72.56} & 57.06 & 40.31 & \textbf{23.26} & 11.40 & \textbf{45.26} & \textbf{3.47} & \textbf{18.80} & \textbf{38.05} & 25.10 & 31.26 & \textbf{35.50} & \textbf{86.68} & 18.11 & \textbf{87.09} & \textbf{28.07} & \textbf{38.90} \\
    \hline
    \end{tabular}
    
\end{table*}




\ifCLASSOPTIONcaptionsoff
  \newpage
\fi



\bibliographystyle{IEEETran}
\bibliography{main}

\end{document}


\title{\textbf{Supplementary Material of ``Learning from 2D: Contrastive Pixel-to-Point Knowledge Transfer for 3D Pretraining''}}
\date{}

\maketitle


\section{Details of ``2D Pretraining on 3D Task''}
In the main paper, we conduct a pilot study to show the possibility of pretraining for 3D scene understanding tasks. Here, we introduce the details of the experiment.

We compare ResNet-50 FCN trained from-scratch or pretrained with ImageNet on ScanNet semantic segmentation \cite{dai2017scannet}. 
The original ScanNet dataset does not provide 2D semantic segmentation labels. In order to acquire pairs of 2D images and segmentation masks, we adopt the technique in \cite{chiang2019unified} and render the semantic label annotated on 3D reconstructed mesh onto the raw RGB images. Therefore, we can train and evaluate the result on 2D using 2D networks. By back-projecting the 2D prediction back to 3D space and aggregate the multi-view prediction in the scene, we can also evaluate the 3D semantic segmentation result using 2D FCN. The result is shown in Table~\ref{tab:scannet-2d3d}. With ImageNet supervised pretraining, 2D FCN performs better in both 2D and 3D ScanNet semantic segmentation mean IoU, which motivates us to pretrain 3D deep neural networks for 3D scene understanding tasks. This also shows that 2D knowledge is able to help 3D scene understanding tasks.
\begin{table}[h]
\begin{center}
\begin{tabular}{l|c|c}
\hline
Method & 2D mIoU & 3D mIoU \\
\hline
From scratch         & 45.38 & 40.46 \\
ImageNet pretrained & 51.13 & 45.27 \\
\hline
\end{tabular}
\end{center}
\caption{\textbf{Evaluation of ImageNet pretrained 2D FCN on 3D ScanNet semantic segmentation}}
\label{tab:scannet-2d3d}
\end{table}


\section{Training Details for Fine-tuning}
\subsection{S3DIS}
We use 5cm voxel size and follow most of the training details in \cite{xie2020pointcontrast}. We use SGD+momentum with learning rate 0.1 and polynomial learning rate scheduler. We apply common 3D data augmentations during fine-tuning, including scaling, rotation, chromatic distortion, elastic distortion, translation, and point dropout. We use batch size 32 and one V100 GPU training for 15k iterations. We report the final validation performance.

\subsection{ScanNet semantic segmentation}
We split the origin training set into train/val 949/525 scenes. We save the best validation score and test on the original validation set.
We use SGD+momentum with learning rate 0.1 and polynomial learning rate scheduler. The batch size is 12 for a single GPU. We use 4 V100 GPUs training for 15k iterations. We use the same augmentations as in S3DIS. 

\subsection{SUN RGB-D and ScanNet object detection}
We use 64 for batch size and 5cm voxel size for SUN RGBD, and batch size 32 and 2.5cm voxel size for ScanNet. We use Adam optimizer with 0.001 learning rate training for 180 epochs on one V100 GPU. The learning rate is decreased 10x after 80 and 120 epochs. For data augmentations, we adopt random rotation, scaling, and flipping. We report the final validation performance.

\section{Baseline Implementation Details}

\subsection{Global KD}
We take the ResNet50 trained by supervised ImageNet classification as the teacher network.
For the student network, we extract the encoder from SR-UNet34 and apply a global mean pooling and a classifier with 1000 output classes. The knowledge distillation loss \cite{hinton2015distilling} is used for training the student by mimicking the output of the teacher before softmax. Unlike general knowledge distillation, we do not apply cross-entropy loss with the ground truth label for the student since there is no labeled data during the pretraining.
\subsection{Global CRD \cite{tian2019contrastiverep} and Gupta \etal~\cite{gupta2016cross}}
Similar to Global KD, we set up the encoder of SR-UNet34 to learn the global output of ImageNet pretrained ResNet50. However, in this case, we do not need a classifier for the 3D network. For Global CRD \cite{tian2019contrastiverep} and Gupta \etal~\cite{gupta2016cross}, we minimize the L2-distance or contrastive loss, respectively, between the global features of the 2D ResNet the 3D encoder.

\section{Experiment Result Details}
This section provide per-class performance for ScanNet, S3DIS, and SUN RGB-D.

\begin{table}[h]
    \footnotesize
    \setlength\tabcolsep{2pt}
    \centering
    \begin{tabular}{l|ccccccccccccc|cc}
    \hline
    Method & clutter & beam & board & book & ceil & chair & column & door & floor & sofa & table & wall & window & mIoU & mAcc \\ \hline
    From scratch & 56.66 & \textbf{0.03} & 72.83 & 68.68 & 92.78 & 89.17 & 31.49 & 68.03 & 98.07 & 58.64 & 76.49 & 82.83 & 51.43 & 65.16 & 73.24 \\
    PointContrast & \textbf{62.01} & 0.00 & 74.39 & \textbf{74.07} & \textbf{94.67} & 90.18 & \textbf{37.35} & 70.33 & \textbf{98.54} & 53.35 & 78.80 & \textbf{84.45} & 51.02 & 66.86 & 73.97 \\
    PPKT & 57.15 & 0.00 & \textbf{78.05} & 73.26 & 93.05 & \textbf{91.00} & 33.31 & \textbf{77.11} & 98.44 & \textbf{70.46} & \textbf{79.77} & 83.67 & \textbf{52.25} & \textbf{68.27} & \textbf{75.19} \\
    \hline
    \end{tabular}
    \caption{\textbf{Per-class IoU on S3DIS semantic segmentation.}}
    \label{tab:scannet-det}
\end{table}

\begin{table}[h]
    \small
    \centering
    \begin{tabular}{l|cccccccccc|c}
    \hline
    Method & bed & table & sofa & chair & toilet & desk & dress & night & book & bath & mAP \\ \hline
    From scratch & 47.35 & \textbf{18.58} & 47.96 & 52.23 & \textbf{61.74} & 5.31 & \textbf{15.65} & 33.16 & 5.69 & \textbf{40.44} & 32.81 \\
    PointContrast & 49.95 & 18.38 & 48.52 & \textbf{53.84} & 53.69 & 5.61 & 14.68 & 36.87 & 7.89 & 37.58 & 32.70 \\
    PPKT & \textbf{52.10} & 17.61 & \textbf{50.34} & 52.31 & 58.05 & \textbf{6.20} & 12.98 & \textbf{43.72} & \textbf{9.36} & 36.55 & \textbf{33.92} \\
    \hline
    \end{tabular}
    \caption{\textbf{Per-class AP under IoU 0.5 on SUN RGB-D object detection.}}
    \label{tab:sunrgbd-det}
\end{table}

\begin{table}[h]
    \scriptsize
    \setlength\tabcolsep{1pt}
    \centering
    \begin{tabular}{l|cccccccccccccccccc|c}
    \hline
    Method & cabin & bed & chair & sofa & table & door & wind & book & pic & cntr & desk & curtn & refrig & shower & toilet & sink & bath & garbg & mAP \\ \hline
    From scratch & 10.61 & 68.40 & 64.86 & 53.54 & 37.69 & 20.64 & 9.80 & 30.08 & 0.64 & 13.92 & 36.63 & 24.70 & \textbf{31.32} & 10.57 & 83.49 & 21.66 & 83.39 & 19.73 & 34.54 \\
    PointContrast & \textbf{12.41} & 67.22 & 70.06 & \textbf{57.91} & \textbf{46.54} & 22.72 & \textbf{11.59} & 37.12 & 0.30 & 10.10 & 36.35 & \textbf{25.47} & 27.30 & 21.64 & 84.96 & \textbf{23.84} & 75.30 & 21.84 & 36.26 \\
    PPKT & 7.95 & \textbf{70.25} & \textbf{72.56} & 57.06 & 40.31 & \textbf{23.26} & 11.40 & \textbf{45.26} & \textbf{3.47} & \textbf{18.80} & \textbf{38.05} & 25.10 & 31.26 & \textbf{35.50} & \textbf{86.68} & 18.11 & \textbf{87.09} & \textbf{28.07} & \textbf{38.90} \\
    \hline
    \end{tabular}
    \caption{\textbf{Per-class AP under IoU 0.5 on ScanNet object detection.}}
    \label{tab:scannet-det}
\end{table}



\begin{algorithm}[t]
\caption{Pseudocode of contrastive pixel-to-point knowledge transfer in Pytorch-like style}\label{alg:ppkt}
\begin{lstlisting}[language=Python]
# f_2D, f_3D: 2D and 3D backbone
# g_2D: upsampling projection layer
# g_3D: projection layer
# bp: back projection function
# n_sub: sub-sample size, nce_t: temperature

for (x, d) in loader:
    (c, f) = bp(x, d)
    z_2D = g_2D(f_2D(x).detach())
    # Back-project pixel-level feature
    _, z_2D = bp(z_2D, d)
    _, z_3D = g_3D(f_3D(c, f))
    
    idxs = random.choice(N, n_sub)
    z_2D = z_2D[idxs]
    z_3D = z_3D[idxs]
    logits = mm(z_3D, z_2D.T)
    labels = arange(n_sub)
    loss = CrossEntropy(logits / nce_t, labels)
    
    loss.backward()
    update(f_3D.param, g_2D.param, g_3D.param)
\end{lstlisting}
\end{algorithm}

{\small
\bibliographystyle{ieee_fullname}
\bibliography{egbib}
}